\documentclass[letterpaper]{article} 
\usepackage{aaai24}  
\usepackage{times}  
\usepackage{helvet}  
\usepackage{courier}  
\usepackage[hyphens]{url}  
\usepackage{graphicx} 
\usepackage{natbib}  
\usepackage{caption} 
\usepackage{algorithm}
\usepackage{algorithmic}
\usepackage{multirow}
\usepackage{algorithm}
\usepackage{algorithmic}
\usepackage{amsmath}
\usepackage{graphicx}
\usepackage{natbib}
\usepackage{tikz}
\usepackage{pgfplots}
\usepackage{newfloat}
\usepackage{listings}
\DeclareCaptionStyle{ruled}{labelfont=normalfont,labelsep=colon,strut=off} 
\floatstyle{ruled}
\newfloat{listing}{tb}{lst}{}
\floatname{listing}{Listing}
\title{Controllable Multi-document Summarization: \\Coverage \& Coherence Intuitive Policy with Large Language Model Based Rewards}
 \author{Litton J Kurisinkel,  Nancy F. Chen \\
         Institute for Infocomm Research, A*STAR, Singapore\\
         litton\_kurisinkel, nfychen@i2r.a-star.edu.sg\\
         }

\usepackage{bibentry}

\begin{document}

\maketitle

\begin{abstract}
Memory-efficient large language models are good at refining text input for better readability. However, controllability is a matter of concern when it comes to text generation tasks with long inputs, such as multi-document summarization. In this work, we investigate for a generic controllable approach for multi-document summarization that leverages the capabilities of LLMs to refine the text. In particular, we train a controllable content extraction scheme to extract the text that will be refined by an LLM. The scheme is designed with a novel coverage and coherence intuitive policy, which is duly rewarded by a passively trained LLM. Our approach yields competitive results in the evaluation using ROUGE metrics and outperforms potential baselines in coherence, as per human evaluation.
\end{abstract}

\section{Introduction}
Relevance of sophisticated multi-document summarization techniques remains unchanged in the era of information explosion. The NLP community has been chasing the problem of multi-document summarization for decades \cite{lin2004rouge}. Earlier techniques for multi-document summarization were based on heuristic text features. Based on these features, they incorporated explicit means to improve topical coverage and diversity of summaries \cite{lin2011class}. They attempted to arrive at a solution using integer linear programming or the greedy method \cite{christensen2013towards}. There were also solutions based on latent semantic features and topic models \cite{ye2016generating}. These techniques were controllable as they operated mainly in discrete space, though they were less capable of learning from the large volume of available training data. Such techniques are almost extinct as the community shifted its focus fully onto data-driven techniques using neural networks \cite{fabbri2019multi}. However, there is a possibility of deriving intuitions from such traditional techniques to improve the controllability of neural multi-document summarization schemes.

Data-driven techniques for summarization using neural networks have been the trend for formulating summarization methods \cite{xiao2021primera}. Data-driven techniques offer several advantages over traditional heuristic-based approaches. Theoretically, they should be capable of automatically learning complex patterns and relationships from data, which can lead to better generalization and adaptability to different types of documents. Additionally, neural networks should be able to capture semantic and syntactic information, resulting in summaries that are more linguistically coherent and fluent. However, neural networks offer fewer provisions to control intermediate computations \cite{alishahi2019analyzing}. In the recent past, large language models based on deep neural networks are capable of producing text that cannot be discriminated from a human-written coherent text \cite{zhao2023more}. However, several memory-efficient large language models have input length constraints when it comes to multi-document summarization \cite{li2023unlocking}. Also, they are not fully exempted from the chance of hallucination \cite{azaria2023internal}.
\begin{figure*}[!t]
  \centering
\includegraphics[width=\textwidth]{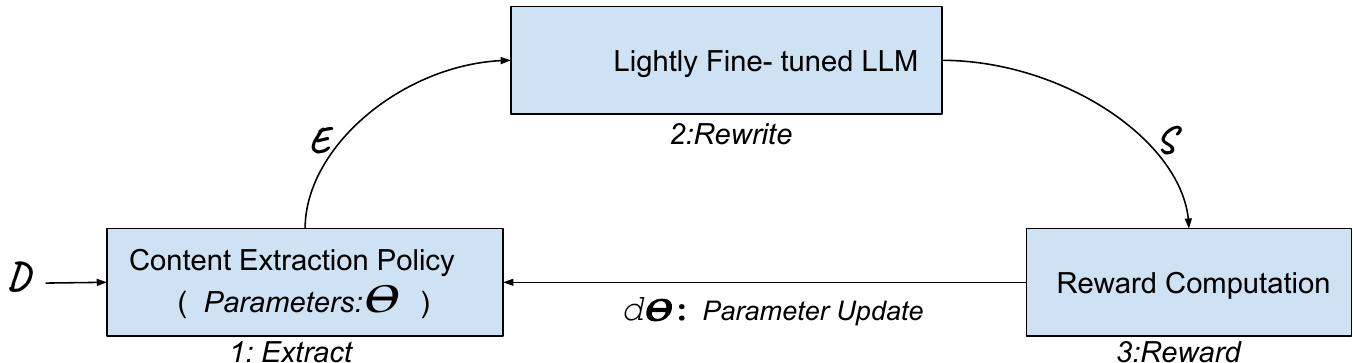} 
  \caption{Extract- Rewrite- Reward Approach for Multi- Document Summarization}
  \label{f:approach}
\end{figure*}

Controllability is a crucial property for any piece of software when it is to be leveraged for practical usage \cite{hu2021causal}. Through the current work, we investigate for an approach that can exhibit the controllability of traditional techniques while being capable of learning from a large amount of training data. Extract-Retrieval approaches retrieve the necessary information and generate the output in a presentable format \cite{liu2020retrieval,lewis2020retrieval}. Such techniques are more tractable and controllable, and there is a possibility to verify retrieved information in comparison with black- box end- to- end generation methods. Inspired by this, we are trying to formulate the problem of multi- document summarization (MDS) using an extract- rewrite approach, which is capable of a joint reinforcement learning. The approach also makes use of the capability of large language models to refine extracted text into a coherent summary. Moreover, such an approach could be scalable for summarizing a larger set of documents \cite{yang2008hierarchical} without being affected by the constraints of smaller context length of memory efficient LLMs \cite{xue2020mt5}. In this context, we introduce a generic framework for MDS with the following components,
\begin{itemize}
    \item A content extraction policy that incorporates explicit means to improve coverage and coherence of the extracted content without the noise of redundant information.
    \item A lightly- trained large language model is used to produce much more coherent content, guided by the extracted text.
    \item A rewarding mechanism which is used to train the extraction policy with respect to the refined text using Reinforcement learning.
\end{itemize}
\section{Previous Works}
 Text summarization can be achieved using extractive methods \cite{lin2011class} and abstractive methods \cite{bing2015abstractive}. Extractive summarization has the advantage of output fluency due to the direct use of human-written texts. However, because sentences exhibit a higher level of granularity regarding the relevant information for the summary, extractive summarizers cannot ensure a noise-free and coherent summary.

 A subset of previous extractive summarization approaches utilized parsed sentence structures to execute noise pruning while extracting content for the summary \cite{morita2013subtree}. As a first step towards abstracting content for summary generation, sentence compression techniques were introduced \cite{lin2003improving}. However, these techniques can merely prune noise and cannot combine related facts from different sentences to generate new ones.

 The attempt to achieve \textit{coherence} in muti-document summarization was attempted by some of the extractive summarization system. \citeauthor{christensen2013towards} \shortcite{christensen2013towards} attempts to achieve structural and topical coherence by a corpus level discourse graph. During summary extraction the system tries to jointly maximize the salience and coherence.
\citeauthor{wang2016exploring} \shortcite{wang2016exploring} try to achieve topical coherence by computing entity role transition probabilities in the corpus. But the attempts to achieve coherence in an extractive summarization scenario often compromises salience for coherence. Also the chance of summary being coherent depends on possibilities existing in the input corpus.

 In certain past attempts, generated summary sentences are merely an optimum recombination of subsentential or phrasal structures \cite{bing2015abstractive}, claiming that the method has the advantage of generating new sentences. \citeauthor{bing2015abstractive} \shortcite{bing2015abstractive} extract relevant noun phrases and verb phrases and recombine them to generate new sentences. 

 Many recent works have developed neural network-based methods for text-to-text generation \cite{Zhong2020extractive, Liu2019fine}. Some of these works focus on generating summaries from input documents \cite{wang2019learning, liu2019text, zhang2020pegasus}. The basic idea is to train a neural network to automatically extract syntactic and semantic features from the input text and generate the desired output. There are extensions of such techniques to a MDS scenario where the input contains more than one document \cite{fabbri2019multi}. In the recent past, there has been a sudden hype in the capability of neural networks to generate text that cannot be distinguished from human-written text, thanks to Large Language Models (LLMs) \cite{sadasivan2023can,gao2023llama, zxhang2023falcon, xue2020mt5}. However, controllability and tractability are matters of concern in many real-life use cases \cite{prabhumoye2020exploring}. Through this current work, we investigate a method that could controllably leverage the capability of LLMs to generate coherent Mult- document summaries. 
\section{Problem Defenition}
We define the problem of controllable multi-document summarization in two steps. 
\begin{align}
 E&= CE(D,\theta) \\
 S&= Z(E)
\end{align}
Where $CE$ is a content extraction method that extracts relevant content $E$ from an input set of documents $D$, which can be refined by a lightly trained content-rewriting model $Z$ into a readable summary $S$. $CE$ should provide provisions to control different attributes of the output summary.
\section{Method}
Our approach for MDS, depicted in Figure \ref{f:approach}, involves major steps
\begin{figure}[!t]
  \centering
  \includegraphics[height=9cm, width=8cm]{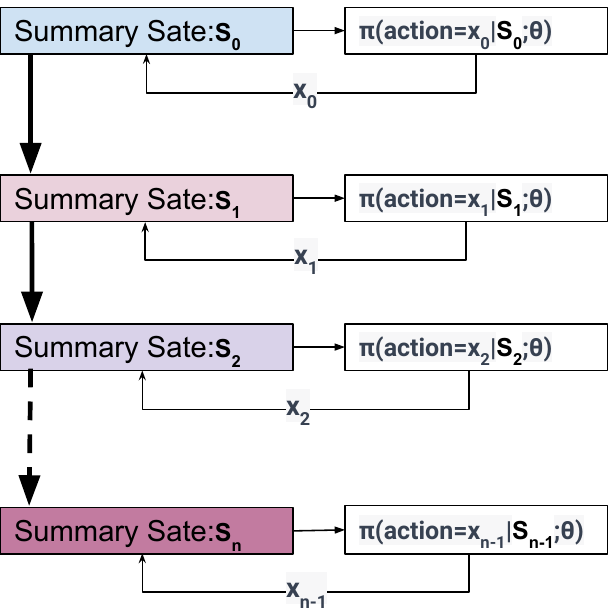} 
  \caption{Multi-Document Summarization in a Reinforcement Learning Setting:
\textbf{States}: Selected Summary Sentences
\textbf{Action}: Selection of the Next Sentence for the Summary Sequence}
  \label{fig:summary_extraction_policy}
\end{figure}
\begin{itemize}
\item \textbf{Extract:} Extract the relevant content from the input set of documents $D$ using a content extraction policy.
\item \textbf{Refine:} Refine the extracted content into readable text using the lightly trained LLM.
\item \textbf{Reward:} Compute the rewards using the refined text to update the parameters of the content extraction policy using the policy gradient method.
\end{itemize}
The rest of this section will explain each one of these sections in detail.
\subsection{Extract}
We aim to extract a coherent sequence of sentences that covers the most relevant content to be included in the summary while minimizing redundant information. From the input set of documents, we extract a trajectory $\tau$ of summary sentences using a content extraction policy $\Pi$. As depicted in Figure \ref{fig:summary_extraction_policy}, at each time-step $t$ of $\tau$, the system's state is represented by the set of already selected sentences $S_{t}$, and the action to be performed is the selection and addition of the next sentence $x_{i}$ to the summary sequence. The execution of an action transitions the summary state from $S_{t}$ to $S_{t+1}$. To achieve this, we formulate a policy that emphasizes content coverage, encourages coherence, and avoids redundancy. At any step $t$ of $\tau$, our policy selects the next sentence $x_{i}$ as follows,
\begin{equation}
\begin{aligned}
z_{t,i}= &cl_{1}*C(x_{i},(D-Set(S_{t}));\theta_{1}) + \\
&cl_{2}*Coh(x_{i},x_{t-1};\theta_{2})
\end{aligned}
\label{control_extraction_eqn}
\end{equation}
\begin{alignat}{2}
\Pi(x_{t}=x_{i}|S_{t}) &= \frac{e^{z_{t,i}}}{\sum_{j=1}^{N-t} e^{z_{t,j}}}
\label{policy_eqn}
\end{alignat}
Where $C$ and $Coh$ are functions that estimate the increase in coverage and coherence values, respectively, with the addition of an argument sentence into the summary. $cl_{1}$ and $cl_{2}$ are control parameters that regulate the coverage and coherence attributes of the output summary. $N$ is the total number of sentences,  $Set(S_{t})$ is the set of sentences already selected and Equation \ref{policy_eqn} computes a probability distribution over the remaining set of sentences $D-Set(S_{t})$. The subsequent subsections will explain $C$ and $Coh$ in detail.
\subsubsection{C:Coverage Function}
The coverage function C estimates how much of the information in the remaining sentences $D-Set(S_{t})$, which are yet to be added to the summary, is covered by $x_{i}$.
\begin{align}
 {\frac{1}{N-t} \cdot \sum_{x_j \in D-Set(S_{t})} F(x_i,x_j)}
\end{align}
Where $F$ is a neural network that takes dense representations $x_{1}$ and $x_{2}$ of argument sentences as input. Subsequently, network computes  $(x_{1}, x_{2})$, $x_{1} * x_{2}$, $x_{1}-x_{2}$, and $|x_{1}-x_{2}|$ \cite{xu2019cross}. These features are concatenated to feed a one-layer MLP with Sigmoid activation to compute the coverage score. The model is made bidirectional by training a forward model with input $(x_{1}, x_{2})$ and a backward model with input $(x_{2}, x_{1})$, both using the same architecture but separate parameters. The coverage score is computed as the average of the two models.
\begin{figure}[!t]
  \centering
  \includegraphics[height=4.5cm, width=8cm]{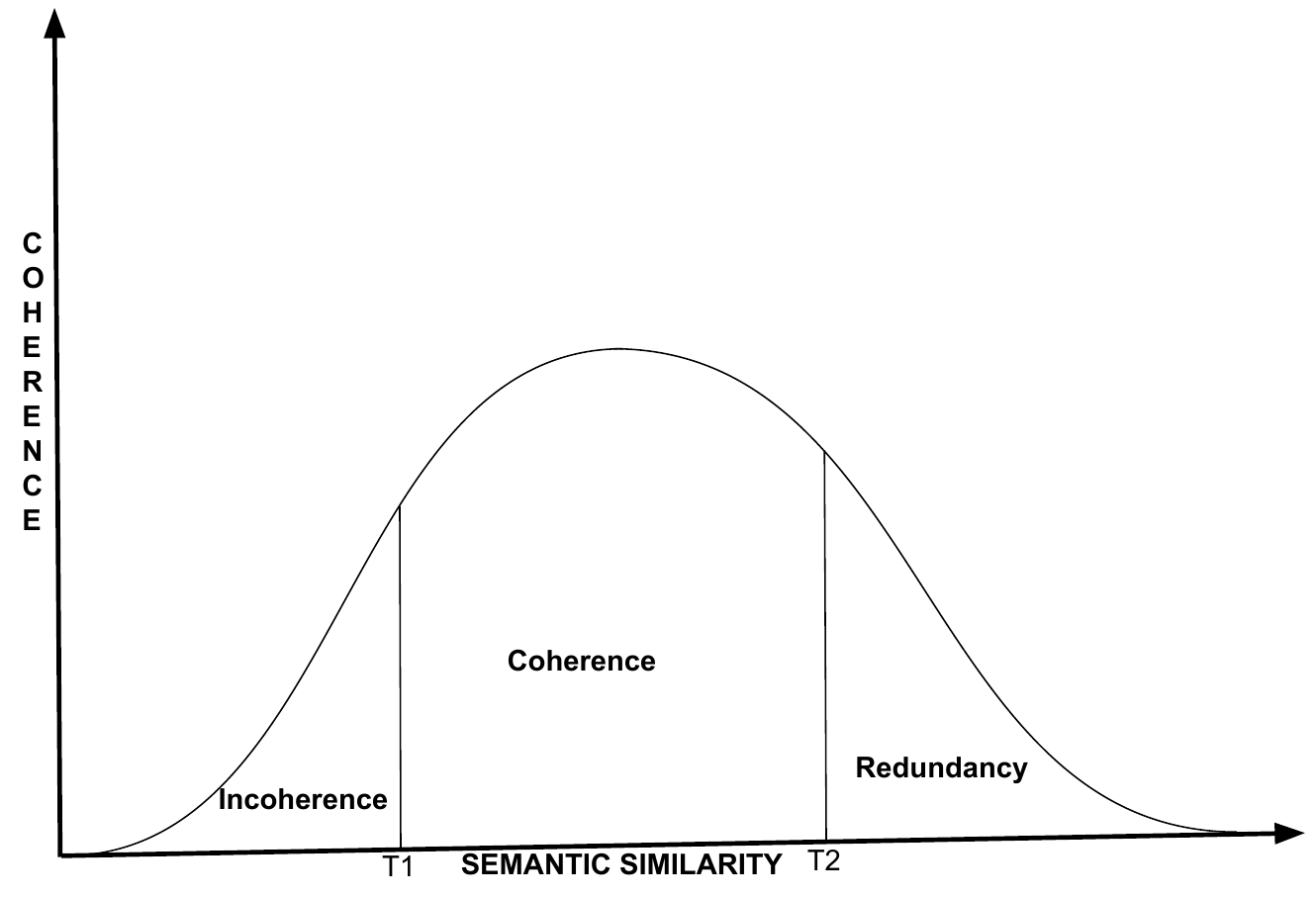} 
  \caption{Trade- off between Coherence and Redundancy \cite{cardenas2022trade} decided by Thresholds T1 and T2}
  \label{fig:cohsim3}
\end{figure}

\subsubsection{Coh: Coherence Score}
In any kind of text generation scenario, the output text should be presented in a coherent manner to ensure lucid reading and quick inference-making by the reader \cite{grosz1995centering}. An incoherent piece of disconnected information won't serve the purpose of a summary. We estimate coherence based on semantic similarity between neighboring sentences in the summary text. As shown in Figure \ref{fig:cohsim3}, for any pair of sentences, if the semantic similarity is below a threshold $T1$, it's considered an incoherent pair. If it is above a threshold $T2$, there will be redundant overlapping information, creating a space of coherence between the thresholds $T1$ and $T2$ \cite{cardenas2022trade}. We relied on the network architecture proposed by \citet{xu2019cross} to estimate the coherence value of two sentences, $x_{1}$ and $x_{2}$. The network computes the vectors $(x_{1}, x_{2})$, $x_{1}*x_{2}$, $x_{1}-x_{2}$, and $|x_{1}-x_{2}|$. The concatenation of these feature vectors is fed into a single-layer MLP to compute the coherence score. The coherence model is pre-trained to identify coherent pairs before being deployed in the reinforcement learning (RL) setting. During pre-training, the loss for a triplet $\mathcal{X_{T}} = (x_{i}, x_{p}, x_{n})$ with positive pair $(x_{1}, x_{p})$ and negative pair $(x_{1}, x_{n})$ is computed as follows,
\begin{align}
\mathcal{L}_{\mathcal{T}} = \max(0, m + Coh(\mathbf{x_{1}}, \mathbf{x_{p}}) - Coh(\mathbf{x_{1}}, \mathbf{x_{n}}))
\label{coherence_loss}
\end{align}
Where $m$ represents the margin. Any pair of consecutive sentences in a human-written summary forms a positive pair $(x_{1}, x_{p})$. A randomly chosen sentence $x_{n}$ from the input set of documents forms a negative pair $(x_{1}, x_{n})$ to identify the threshold $T1$. When a sentence is paired with itself, it forms a negative pair $(x_{1}, x_{1})$ to identify the threshold $T2$.
\subsubsection{Number of Sentences:}
The number of sentences to be extracted, denoted as $TN$, is determined based on the variance $\sigma^2$ among the similarity values between input sentences. This similarity is measured as the cosine similarity between the corresponding sentence representations.
\begin{align}
TN = \lfloor k + c \cdot \sigma^2 \rfloor \label{eq:number-of-sentences}
\end{align}
Where $k$ and $c$ are constants that are optimized through grid search. Greater the variance, more the number of sentences.
\subsection{Z: Re- writing Model}
The re-writing model takes the extracted sentences and rewrites them into a coherent and readable summary. During this process, the model should be capable of ordering or aggregating information and adding discourse markers if necessary. We trained such a model using a memory-efficient large language model for computational efficiency during both training and inference. To perform fine-tuning on the large language model (LLM), we aligned summary sentences in different summarization datasets with source sentences in the input set of documents \cite{fabbri2019multi,ghalandari2020large}. This alignment was achieved using the method described by \citet{wolhandler2022multi}. Specifically, we fine-tuned the flan-t5-xl model\footnote{https://huggingface.co/google/flan-t5-xl}, which is designed for text-to-text generation, and introduced the 're-write' prompt during training. The flan-t5-xl model is relatively lightweight and requires less computational resources for both training and inference.
\subsection{Reward Computation}
Once the rewritten summary is ready, we need to reward the content extraction process by comparing the rewritten coherent summary with the reference summary. The comparison is based on semantic similarity and N-gram overlap. Based on this rewarding scheme, we compute the training loss for our extractive mechanism as follows.
\begin{align}
\begin{split}
\mathcal{L_{\tau}} &= -R_{\tau}\sum_{t=0}^{TN}  \cdot \log(\pi(x_t|s_t)) \\
&\quad - \lambda \frac{1}{TN} \sum_{t=0}^{TN} (\pi(x_t|s_t) - {r_{t}})^2
\end{split}
\label{equation_loss}
\end{align}

Where $R_{\tau}$ is the cumulative reward, computed as follows,
\begin{align}
R_{\tau} &= \frac{1}{2}(\text{ROUGE}(S_{final},S_{\text{ref}}) + \text{Sim}(S_{final},S_{\text{ref}}))
\end{align}
Where ROUGE computes the average of ROUGE-L and ROUGE-2 scores for the final rewritten summary $S_{final}$ with respect to the reference summary $S_{ref}$, while Sim computes the cosine similarity between the S-BERT \footnote{https://huggingface.co/Muennighoff/SBERT-base-nli-v2} representations of $S_{final}$ and $S_{ref}$. The reward $r_{t}$ for selecting an individual sentence $x_{t}$ at time-step $t$ is computed as follows: 
\begin{align}
r_{t} &= \frac{1}{2}(\text{ROUGE}(x_{t},S_{\text{ref}}) + \text{Sim}(x_{t},S_{\text{ref}}))
\end{align}
The second term in Equation \ref{equation_loss} is inspired by the actor-critic method \cite{fujimoto2018addressing}. However, we utilize the action prediction probability directly instead of a value function.
\begin{table}
    \centering
    \label{tab:metrics}
    \begin{tabular}{|c|c|c|}
        \hline
        \textbf{Precision} & \textbf{Recall} & \textbf{F-Measure} \\
        \hline
        1.0 & 0.70 & 0.82 \\
        \hline
    \end{tabular}
    \caption{Coherence Estimation Model Pre- training Results}
    \label{coherence_results}
\end{table}
\begin{table*}[t]
  \centering
  \begin{tabular}{lcccc}
    \hline
    \textbf{Dataset} & \textbf{Model} & \textbf{ROUGE-1} & \textbf{ROUGE-2} & \textbf{ROUGE-L} \\
    \hline
    \multirow{6}{*}{Multi- News} & HiMAP &  44.17 & 16.05 & 21.38 \\
    & Hierarchical Transformer  &  42.36 &  15.27 & 22.08 \\
    & GraphSum & 45.02 & 16.69 & 22.50 \\
    & GraphSum + RoBERTa &  45.87 & 17.56 & 23.39 \\
    & BART-Long & \textbf{48.54} & \textbf{18.56} &  \textbf{23.78} \\
    & Current Method& 46.27 & 18.0 & 24.30 \\
    & Current Method + RL& 46.50 & 18.18 & 24.73\\
    \hline
    \multirow{6}{*}{WCEP} & TSR & 35.30 &  13.70 & 25.70 \\
    & BERTReg & 35.00 & 13.50 & 25.50 \\
    & BART-WCEP-DynE-5 &  35.40 & 15.10 & 25.60 \\
    & Current method & 39.45 & 17.90 & 30.26 \\
    & Current method + RL & \textbf{40.71} &\textbf{18.34} & \textbf{31.58}\\
    \hline
  \end{tabular}
  \caption{Text Summarization Results with ROUGE Metrics}
  \label{tab:rouge_results}
\end{table*}
\begin{table}
\centering
\begin{tabular}{lccc}
\hline
\textbf{Model} & \textbf{ROUGE-1} & \textbf{ROUGE-2} & \textbf{ROUGE-L} \\
\hline
Lin+Z &  36.10 & 14.10 & 25.00\\
G-Flow+Z & 34.30 & 12.00 & 22.30 \\
CM + RL & \textbf{40.71} &\textbf{18.34} &  \textbf{31.58}\\
\hline
\end{tabular}
\caption{Comparison with Extract- Rewrite Models: CM denotes $Current Method$}
\label{tab:extractive-rewrite}
\end{table}
\section{Experiments and Results}
We conducted experiments to evaluate summaries generated with different attributes, namely content coverage and coherence. We relied on objective metrics to evaluate content coverage, while human evaluation was employed for assessing coherence. Additionally, we conducted experiments to evaluate our coherence estimation and re-writing sub-models.
\subsection{Sub Models}
\subsubsection{Coherence Model}
To train the coherence estimation submodel, we gathered summary texts from the MultiNews and WCEP datasets to create training sets. These positive and negative pairs were selected to construct the triplet in Equation \ref{coherence_loss}. Our dataset comprised 45,000 training records, in addition to 2,700 development records and 2,700 testing records, respectively. The results of the model's performance in identifying coherent pairs are presented in Table \ref{coherence_results}. The model achieved a reliable F-Measure score of 0.82.
\subsubsection{Fine-tuning LLM for Re- writing} 
We utilized the flan-t5-xl model\footnote{https://huggingface.co/google/flan-t5-xl} for the purpose of re-writing. As previously explained in the paper, we constructed a dataset by aligning summary sentences from the Multi-news dataset with source sentences. This process resulted in approximately 3000 parallel records for fine-tuning, 450 for development, and 450 for testing. The fine-tuned re-writing model yielded a BLEU score of 0.30.
\subsection{Summarization}
We conducted experiments to evaluate our summaries on the dimensions of content coverage and coherence. We relied on objective metrics such as ROUGE-1, ROUGE-2, and ROUGE-L \cite{lin2004rouge} to estimate content coverage, while human evaluation was employed to estimate coherence. 
\subsubsection{Data:}
We assessed our summarization approach using public datasets, including Multi-news \cite{fabbri2019multi} and WCEP test sets \cite{ghalandari2020large}. Multi-news contains clusters of related news documents as input, along with their corresponding auto-aligned summaries. The WCEP dataset for multi-document summarization (MDS) comprises short, human-written summaries about news events, sourced from the Wikipedia Current Events Portal (WCEP). Each summary is paired with a cluster of news articles associated with a specific event.
\subsubsection{Settings:}
We utilized the S-BERT model\footnote{https://huggingface.co/Muennighoff/SBERT-base-nli-v2} to compute sentence representations. Our summarization model was developed in two stages. In the pre-training phase, the learning rates of sentence representations and the already trained coherence submodel were set to zero. The coverage function in the network was learned as a regression model using the second component in the loss function (Equation \ref{equation_loss}). The learning rate of the network was set to $10^{-4}$ during pre-training. During the subsequent training phase, the entire network was trained using reinforcement learning, incorporating both components in the loss function (Equation \ref{equation_loss}). At this stage, the learning rate was adjusted to $10^{-6}$. For each dataset, constants $cl_{1}$, $cl_{2}$, $k$, $c$ and $\lambda$ are optimized using development set to obtain maximum ROUGE-2 + ROUGE-L score.
\subsubsection{Results: Content Coverage}
We estimated the content coverage of summaries using the objective metrics ROUGE-1, ROUGE-2, and ROUGE-L. In the evaluation using the multi-news dataset, we compared our model with peer systems such as HiMAP \cite{fabbri2019multi}, Hierarchical Transformer \cite{liu2019hierarchical}, GraphSum \cite{li2020leveraging} and BART-Long 
 \cite{pasunuru2021efficiently}. We also compared the reported results in WCEP datasets of systems such as TSR \cite{ghalandari2020large}, BERTReg \cite{ghalandari2020large}, and BART-WCEP-DynE-5 \cite{hokamp2020dyne}. Results are shown in Table \ref{tab:rouge_results}. The $Current method$ serves as our pre-trained model, while the augmentation of the pre-trained model with reinforcement learning is denoted as $Current method + RL$. Our approach yields results that are comparable with peer systems in general and beats a few potential baselines. We observe that incorporating explicit means to ensure information coverage in the model helped in ensuring competitive ROUGE scores while keeping the model controllable. Through the current work, we investigated a controllable method for multi-document summarization based on an extract-rewrite approach. So, it is essential to compare with other extract-rewrite methods possible. For this purpose, we leveraged extractive MDS approaches such as \cite{lin2011class} and \cite{christensen2013towards} to create extractive summaries. The number of sentences to be extracted is computed using Equation \ref{eq:number-of-sentences}. Later, we rewrite these summaries using our rewriting model $Z$. We name these settings $Lin+Z$ and  $G-Flow+Z$ respectively. The results are shown in Table \ref{tab:extractive-rewrite}. Our approach outperforms baselines in terms of all three metrics considered. This indicates that a trainable content extraction policy can improve content coverage in a controllable extract-rewrite approach.
\begin{figure}[!t]
  \centering
  \includegraphics[height=5.4cm, width=9cm]{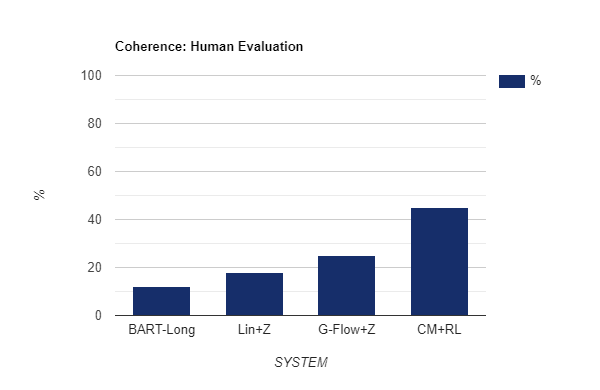} 
  \caption{Human Evaluation for Coherence: Y axis of the graph represents the percentage of times each system is chosen by the evaluators during the experiment. CM denotes current method.}
  \label{fig:coh_eval}
\end{figure}
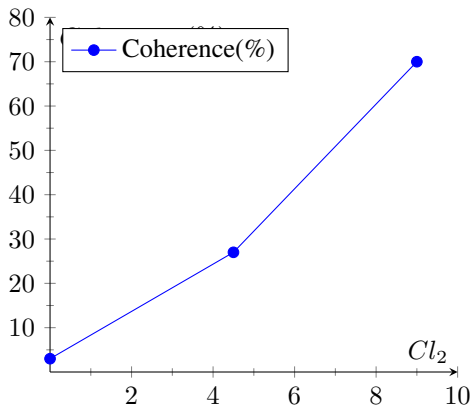
\begin{figure}
    \centering
    \begin{tikzpicture}
    \begin{axis}[
        xlabel={$Cl_{2}$},
        ylabel={$Coherence(\%)$},
        xmin=0, xmax=10,
        ymin=0, ymax=80,
        xtick={0, 2, ..., 10},
        ytick={0, 10, ..., 80},
        grid=none,
        major grid style={line width=.2pt,draw=gray!50},
        minor tick num=1,
        width=7cm, 
        height=6.3cm, 
        axis lines=middle,
        legend pos=north west,
    ]
    
    \addplot[color=blue,mark=*] coordinates {
        (0,3)
        (4.5,27)
        (9,70)
    };
    \legend{Coherence(\%)}
    
    \end{axis}
    \end{tikzpicture}
    \caption{Controllability: Correlation of Coherence with control parameter $cl_{2}$}
    \label{fig:control_cl2}
\end{figure}
\subsection{Human Evaluation for Coherence}
To evaluate coherence, we chose four human evaluators who are postgraduate students in linguistics. We randomly selected a sample set of output summaries consisting of 45 summary sets from different datasets. Each summary set contains summaries generated by the settings, namely $Lin+Z$, $G-Flow+Z$ and $Current Method+RL$. The summaries are shown to the evaluators in a random order to avoid any kind of bias and they are asked to choose the most coherent among the listed summaries. The evaluators are instructed to estimate coherence based on discourse connections using linguistics cues and topical continuity between neighbouring sentences in the summaries. The results are shown as graph in the Figure \ref{fig:coh_eval}. Y axis of the graph represents the percentage of times each system is chosen by the evaluators during the experiment. $Current Method+RL$ is overwhelmingly selected by the evaluators in comparison with $Lin+Z$ and $BART- Long$. We observe that $current Method+RL$ employed explicit provisions  to extract a coherent sequence of sentences during content extraction. As a consequence of this extraction process worked as planning mechanism during rewriting using $Z$.   Even after incorporating means to ensure coherence, $G-Flow+Z$ didn't perform well during human evaluation. They used crude discrete methods to estimate coherence which can't be trained with the datasets. Modern day neural networks are equipped to generate more coherent text. We pre- trained our coherence model  using  human-written coherent text and did minimal fine- tuning during RL for summarization. We also conducted human evaluations to assess the controllability of coherence as a summary attribute. We generated summaries for different values of $cl_{2}$, the control parameter for coherence in Equation \ref{control_extraction_eqn}, for each of the 45 document sets. We then repeated the human evaluation for coherence using summary sets generated. The results are depicted in Figure \ref{fig:control_cl2}. It is evident from the results that increasing $cl_{2}$ enhances coherence.
The approach involves two separate components for content extraction and re-writing, which can be deployed independently. The deployed components can be accessed via separate web services sequentially to achieve the final output.
\section{Conclusion}
In our current work, we introduced an approach that offers control over multi-document summarization. This approach demonstrates superior performance compared to potential baseline methods, as evidenced by objective evaluations using ROUGE metrics. Furthermore, the method's effectiveness is underscored by human evaluators, who found improved coherence in the generated summaries. We have also investigated and validated the controllability of coherence by adjusting the associated control parameter. Notably, this approach could be adaptable across various domains, serving as a generic framework. For instance, it can facilitate the summarization of patient records through the use of a Large Language Model (LLM), where clinically significant information is given more reward during the training of the extraction policy.
\bibliography{aaai24}
\end{document}